\documentclass[10pt,twocolumn,letterpaper]{article}

\usepackage[pagenumbers]{cvpr} 

\usepackage[dvipsnames]{xcolor}

\definecolor{cvprblue}{rgb}{0.21,0.49,0.74}
\usepackage[pagebackref,breaklinks,colorlinks,citecolor=cvprblue]{hyperref}
\usepackage{multirow}

\usepackage[accsupp]{axessibility} 

\def\paperID{46} 
\def\confName{5th Workshop on Vision for Agriculture}
\def\confYear{2024}

\title{Gasformer: A Transformer-based Architecture for Segmenting Methane Emissions from Livestock in Optical Gas Imaging}

\author{Toqi Tahamid Sarker\textsuperscript{1}, Mohamed G Embaby\textsuperscript{2}, Khaled R Ahmed\textsuperscript{1}, Amer AbuGhazaleh\textsuperscript{2}\\
\textsuperscript{1}School of Computing, \textsuperscript{2}School of Agricultural Sciences\\
Southern Illinois University, Carbondale, USA\\
{\tt\small \{toqitahamid.sarker, mohamed.embaby, khaled.ahmed, aabugha\}@siu.edu}
}

\begin{document}
\maketitle
\begin{abstract}
Methane emissions from livestock, particularly cattle, significantly contribute to climate change. Effective methane emission mitigation strategies are crucial as the global population and demand for livestock products increase. We introduce Gasformer, a novel semantic segmentation architecture for detecting low-flow rate methane emissions from livestock, and controlled release experiments using optical gas imaging. We present two unique datasets captured with a FLIR GF77 OGI camera. Gasformer leverages a Mix Vision Transformer encoder and a Light-Ham decoder to generate multi-scale features and refine segmentation maps. Gasformer outperforms other state-of-the-art models on both datasets, demonstrating its effectiveness in detecting and segmenting methane plumes in controlled and real-world scenarios. On the livestock dataset, Gasformer achieves mIoU of 88.56\%, surpassing other state-of-the-art models.
Materials are available at:
\href{https://github.com/toqitahamid/Gasformer}{\color{blue}{$\tt github.com/toqitahamid/Gasformer$}}.
\end{abstract}

\section{Introduction}

Methane, a potent greenhouse gas, significantly contributes to climate change and global warming. Although methane has a shorter atmospheric lifetime than carbon dioxide, it has a much higher global warming potential, being 36 times more potent per kg than CO2 over a 100-year period \cite{Us_Epa2016-vj}. Human activities, such as agriculture, waste disposal, and fossil fuel production, are responsible for about 60\% of global methane emissions, accounting for at least 25\% of global warming \cite{Environment2021-fo}. The livestock sector, particularly ruminant animals like cattle, is a significant source of methane emissions due to enteric fermentation from their digestive process and manure management practices \cite{gleam}. Ruminant livestock account for about 6\% of total global greenhouse gas emissions \cite{Kirschke2013-uw,Beauchemin2020-mr}. In addition to the negative climate impacts, enteric methane release also represents a loss of up to 15\% of the dietary energy fed to the animals, hindering production efficiency \cite{holter1992methane}.

The world's population is expected to grow significantly, reaching around 8.5 billion by 2030, 9.7 billion by 2050, and 10.9 billion by 2100 \cite{United_Nations_Department_of_Economic_and_Social_Affairs_Population_Division2019-fa}. This population growth is the primary driver of increasing demand for livestock products, with estimates suggesting that livestock production needs to increase by 70\% over the next 50 years to meet the rising global demands \cite{Rojas-Downing2017-pi}. Consequently, as the global population and demand for livestock products increase, methane emissions will continue to rise unless effective mitigation strategies are developed.

Recent research has focused on accurately detecting and measuring methane emissions from livestock operations and other agricultural sources to develop mitigation strategies. Various approaches have been explored to estimate methane emissions, aiming to improve efficiency and accuracy compared to traditional manual surveys and inspections. Jeong et al. \cite{Jeong2022-ga} employed artificial intelligence techniques to estimate methane emissions from dairy farms in California's San Joaquin Valley, using a U-Net model to identify dairy facilities from aerial images and calculate their spatial area. Using the facility area, they estimated the number of dairy cows and their methane emissions from digestion and manure. Ramirez-Agudelo et al. \cite{ramirez2022intake} proposed the use of computer vision to quantify intake time as a predictor of methane emissions from cattle, applying the YOLOv5x object detection model on video footage. While these studies have significantly contributed to methane emission estimation, they have primarily focused on indirect estimation techniques rather than direct detection and visualization of methane emissions.

Optical Gas Imaging (OGI) is a powerful technology that has revolutionized the detection and visualization of gas leaks, particularly methane. Several studies have utilized OGI cameras for methane leak rate classification, such as Wang et al. \cite{Wang2020-em} using an OGI camera to classify methane leaks from a controlled-release test facility that mimics real-world gas leaks found at natural gas production sites and Wang et al. \cite{Wang2022-mj} exploring more advanced architectures like 3D CNN and ConvLSTM. However, these studies focused on high methane flow rates when capturing the gas with OGI cameras, and there is a need for research that explores advanced computer vision and imaging techniques to directly detect and visualize methane emissions from livestock operations and other agricultural sources in real-time, particularly at lower flow rates associated with cattle emissions.

To accurately detect and quantify methane emissions at lower flow rates, such as those associated with cattle emissions, advanced image segmentation techniques are needed. Semantic segmentation, a computer vision task that assigns a class label to each pixel in an image, has been applied to various problems, such as medical images \cite{Ronneberger2015-yu} and aerial images \cite{Marmanis2016-we}. Fully Convolutional Networks (FCNs) \cite{fcn} revolutionized this field by enabling end-to-end training of segmentation models. However, transformer \cite{Vaswani2017-ju} models have now emerged as the dominant architecture since the introduction of Vision Transformer (ViT) \cite{Dosovitskiy2020-gq}, outperforming traditional convolutional neural networks (CNNs) in tasks like image classification, object detection, and image segmentation. Applying semantic segmentation techniques to OGI images of methane emissions can help accurately detect and quantify methane plumes at lower flow rates, providing valuable insights for developing effective mitigation strategies.

In this paper, we focus on segmenting methane plumes with low flow rates using semantic segmentation models. We have created two unique datasets using the FLIR GF77 OGI camera: the Controlled Methane Release (MR) Dataset for methane segmentation and quantification and the Dairy Cow Rumen Gas Dataset (CR) for methane segmentation. The first dataset is collected using a flow rate controller connected to a methane cylinder, allowing for the controlled release of methane at specific flow rates. Controlled gas release is crucial for understanding the relationship between methane flow rate and the corresponding OGI camera response. The second dataset was collected from dairy cow rumen gas samples obtained from a farm. By collecting OGI camera footage of methane emissions directly from dairy cow rumen gas, this dataset provides a more realistic representation of cattle methane emissions.

We propose a novel architecture called Gasformer for semantic segmentation of methane gas in OGI camera images. In the encoding stage, we incorporate Mix Vision Transformer \cite{Xie2021-in}, designed to generate multi-scale features from an input image without the need for positional encoding. This approach allows for the efficient extraction of relevant features at different scales, which is essential for accurately segmenting methane plumes of varying sizes and shapes. In the decoding stage, the Light-Ham \cite{lightham} decoder takes multiple feature maps from different levels of the encoder as input and applies a Hamburger \cite{Geng2021-mg} module consisting of matrix decomposition sandwiched between linear transformations to refine feature maps for semantic segmentation. The main contributions of this study are as follows:

\begin{enumerate}
    \item We propose Gasformer, a novel segmentation architecture for methane gas detection and quantification.
    \item We introduce two new datasets with corresponding labels for segmenting low-flow rate methane gas and dairy cow rumen gas.

    \item We evaluate Gasformer's performance on both datasets and compare it with state-of-the-art segmentation models.
\end{enumerate}


\section{Methods}

\begin{figure*}[!t]
        \centering
        \small
        \resizebox{0.70\linewidth}{!}{%
        \includegraphics[width=1\linewidth]{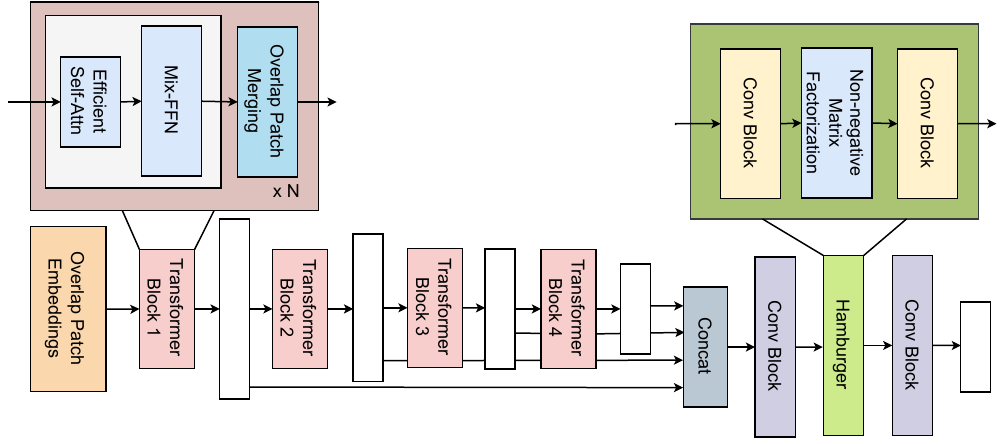}}
        \vspace{-8pt}
        \caption{Overall framework of Gasformer. We extract multi-scale features from the encoders, and the decoder concatenates these multi-level features and predicts the segmentation mask }
         \label{fig:gasformer}%
        \vspace{-8pt}
\end{figure*}

Our proposed model is based on encode-decoder architecture. The overall architecture of the model is shown in \Cref{fig:gasformer}.


\subsection{Encoder}

The Gasformer encoder is based on the Mix Vision Transformer (MiT-B0) \cite{Xie2021-in} architecture, which generates multi-scale feature representations from the input image without explicit positional encoding. The input RGB images captured by an Optical Gas Imaging (OGI) camera are divided into overlapping patches using the Overlap Patch Embedding operation. In the first stage, patches of size 7 × 7 are extracted with a stride of 4 and padding of 3, resulting in a feature map with 1/4 the spatial resolution of the input image.
The patch embeddings are processed by a hierarchical stack of Transformer blocks, each operating at a different spatial resolution. The Transformer blocks consist of three main components: Efficient Self-Attention, Mix-FFN, and Overlap Patch Merging.

\noindent\textbf{Efficient Self-Attention.} The Efficient Self-Attention mechanism in the Transformer block computes the attention weights between patches using a multi-head self-attention operation. It projects the patch embeddings into query (Q), key (K), and value (V) matrices. The attention weights are obtained by computing the scaled dot-product between the query and key matrices, followed by a softmax function:

\vspace{-5pt}
\begin{align}
\operatorname{Attention}(Q, K, V)=\operatorname{Softmax}\left(\frac{Q K^{\top}}{\sqrt{d}}\right) V
\end{align}
\vspace{-5pt}

Where d is the dimension of the query and key vectors, and $\sqrt{d}$ is a scaling factor that helps prevent the attention weights from becoming too small or too large. The resulting attention matrix has dimensions of $n$ × $n$, where each element represents the similarity between a pair of patches.

To reduce the computational complexity of the self-attention operation, the Gasformer encoder employs a sequence reduction process, as described in \cite{Wang2021-xo}. This process reduces the spatial dimensions of $K$ by a reduction ratio r, effectively reducing the self-attention computation from $O(n^2)$ to $O(n^2/r)$. 

\noindent\textbf{Mix-FFN.} The Mix-FFN block replaces the traditional positional encoding by incorporating a 3 × 3 convolution into the feed-forward network (FFN). The Mix-FFN block consists of a multilayer perception (MLP), followed by a 3 × 3 depth-wise convolution, GELU activation, and another MLP with a residual connection. The depth-wise convolution applies the convolution operation independently on each channel of the input feature map, helping to capture channel-specific spatial relationships while being computationally efficient. By leveraging the inherent property of convolutions to leak location information through zero padding \cite{Islam2020-lv}, the Mix-FFN block encodes local spatial relationships directly into the feature representations.

\noindent\textbf{Overlap Patch Merging.} The Overlap Patch Merging operation downsamples the feature maps and generates multi-scale representations. It is applied after each Transformer block stage, reducing the spatial resolution while increasing the channel dimension. The subsequent stages use patch sizes of 3 × 3, stride of 2, and padding of 1, generating feature maps with resolutions of 1/8, 1/16, and 1/32 of the input image size.
The Gasformer encoder generates four multi-scale feature maps, denoted as F1, F2, F3, and F4, with resolutions of 1/4, 1/8, 1/16, and 1/32 of the input image size, respectively. These multi-scale features capture both local and global context, which is essential for accurate methane plume segmentation.

\subsection{Decoder}

In this study, we adopt Light-Ham \cite{lightham}, a lightweight and efficient decoder architecture for semantic segmentation. 
The Light-Ham decoder takes multiple feature maps from the encoder as input and concatenates the features from all stages of the encoder. It then employs a lightweight Hamburger \cite{Geng2021-mg} module to further model the global context. The key component of the decoder is the Hamburger Matrix Decomposition (HamMD) module, which efficiently captures long-range dependencies by performing matrix decomposition using Non-negative Matrix Factorization (NMF). This allows for an efficient representation of the input features.

The architecture of Light-Ham consists of three main components: a squeeze layer, the HamMD module, and an alignment layer. The squeeze layer reduces the channel dimensions of the input feature maps, while the HamMD module captures long-range dependencies. The alignment layer then matches the output dimensions of the HamMD module to the desired output size.

\section{Dataset}


\subsection{FLIR GF77 Optical Gas Imaging Camera}
\label{sec:flir}
In this study, we use a FLIR GF77 Gas Imaging Camera with a GF77-LR low-range lens to capture images of methane plumes. The FLIR GF77 is an uncooled long-wave infrared (LWIR) camera, making it suitable for operation at any temperature, although it is less sensitive than cooled cameras for detecting methane gas. The camera's ability to detect methane is determined by the Noise Equivalent Concentration Length (NECL), which quantifies the minimum detectable gas concentration over a given path length. For the FLIR GF77 camera, the NECL gas sensitivity for methane is 100 ppm-m, meaning it can detect methane plumes with a concentration of at least 100 parts per million over a path length of one meter. The camera operates within a spectral range of $7\,\mu\text{m}$ to $8.5\,\mu\text{m}$.

\begin{figure}
    \centering
    \rotatebox[origin=c]{90}{\parbox{0cm}{\centering RGB}}\hspace{5pt}%
    \subfloat{\includegraphics[width=0.25\columnwidth]{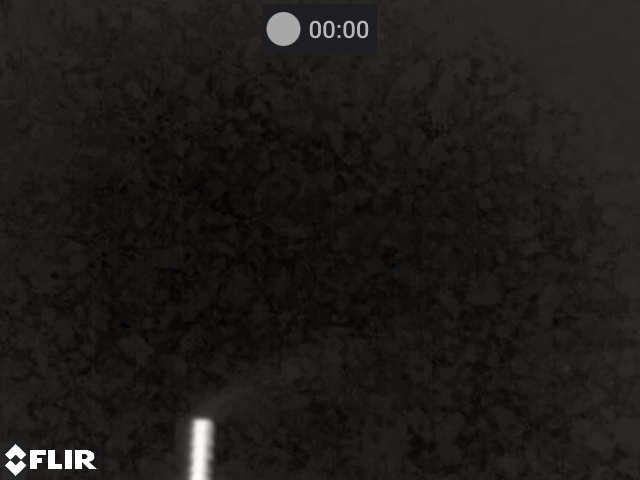}} \hspace{0pt}
    \subfloat{\includegraphics[width=0.25\columnwidth]{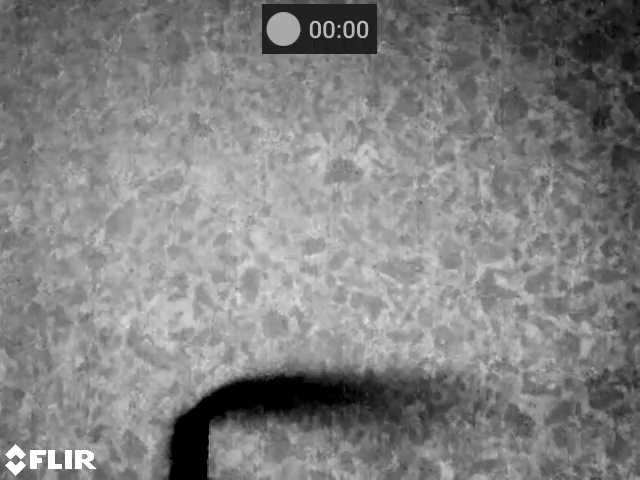}} \hspace{0pt}
    \subfloat{\includegraphics[width=0.25\columnwidth]{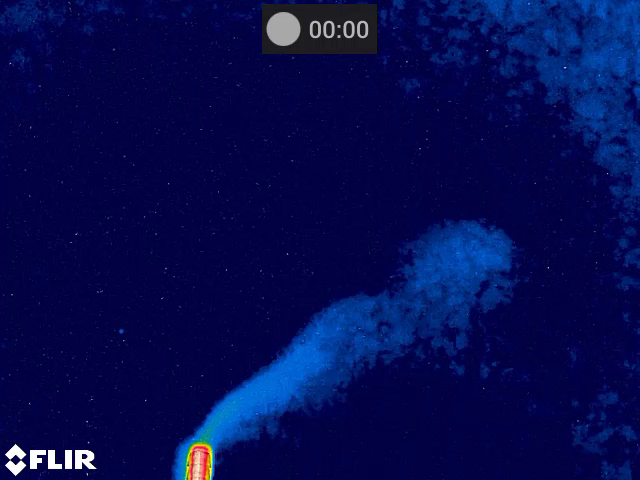}}  \\
    \setcounter{subfigure}{0}
    \rotatebox[origin=c]{90}{\parbox{0cm}{\centering Mask}}\hspace{5pt}%
    \subfloat[10 SCCM]{\includegraphics[width=0.25\columnwidth]{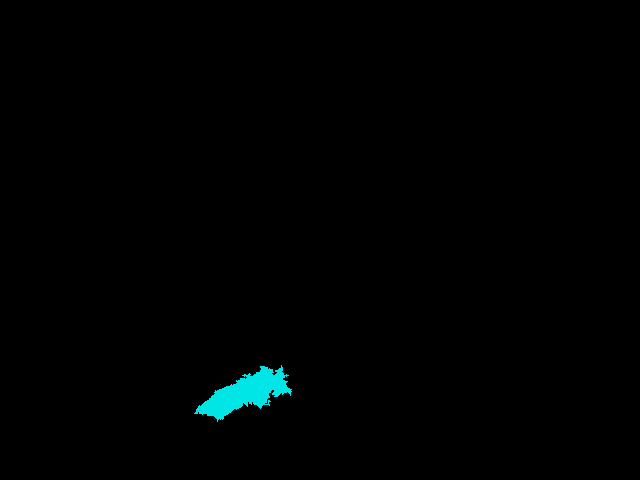}} \hspace{0pt}
    \subfloat[60 SCCM]{\includegraphics[width=0.25\columnwidth]{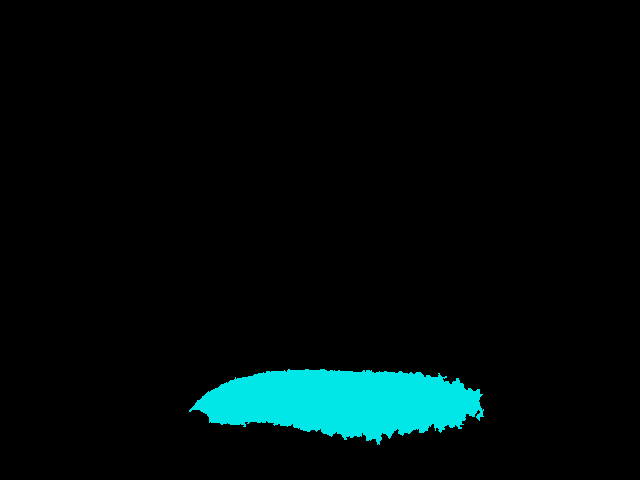}} \hspace{0pt}
    \subfloat[80 SCCM]{\includegraphics[width=0.25\columnwidth]{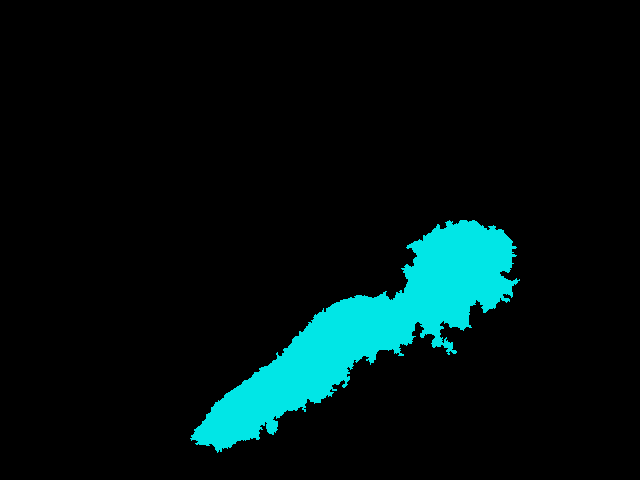}} \\
    \vspace{-5pt}
    \caption{Controlled methane release visualizations at varying flow rates on MR dataset. (a) Flow rate of 10 standard cubic centimeters per minute (SCCM). (b) Flow rate of 60 SCCM. (c) Flow rate of 80 SCCM.}
    \label{fig:dataset_methane}
    \vspace{-5pt}
\end{figure}


To successfully detect methane gas using the FLIR GF77 OGI camera, four critical factors must be considered:

\vspace{2pt}
\noindent\textbf{Response Factor.} The infrared absorption spectrum of methane must align with the camera's spectral sensitivity range. Methane has a response factor of 0.297, indicating that its distinct infrared absorption bands, including a major band centered around $7.7 \pm 0.1\,\mu\text{m}$, overlap well with the FLIR GF77's operational spectral range. This overlap allows the camera's uncooled microbolometer detector, with a 320 x 240 pixel resolution and a thermal sensitivity of $<$25 mK, expressed as Noise Equivalent Temperature Difference (NETD), to effectively detect and visualize methane gas plumes by measuring the absorbed infrared radiation within the shared wavelength region.

\vspace{2pt}
\noindent\textbf{Temperature Difference.} A noticeable temperature difference of at least 3\textdegree C must exist between the methane plume and the surrounding background environment for the FLIR GF77 to capture the gas effectively.

\vspace{2pt}
\noindent\textbf{Gas Concentration.} The concentration or flow rate of methane present in the captured image must exceed the camera's minimum detectable limit, defined by the NECL gas sensitivity. For successful visualization, the gas concentration and path length must surpass the FLIR GF77's sensitivity threshold of at least 100 ppm-m for methane when the temperature difference between the plume and background is 10\textdegree C and the distance is 1 meter.

\vspace{2pt}
\noindent\textbf{Distance.} The distance from the camera to the methane plume also plays a role in the ability to detect and visualize the gas effectively.

When methane is present in the camera's field of view and meets these criteria, it will absorb and attenuate approximately 29.7\% of the infrared radiation within the shared $7\,\mu\text{m}$ to $8.5\,\mu\text{m}$ range, creating a contrast against the background that the FLIR GF77 can capture and display as an image.

\begin{figure}
    \centering
    \subfloat{\includegraphics[width=0.25\columnwidth]{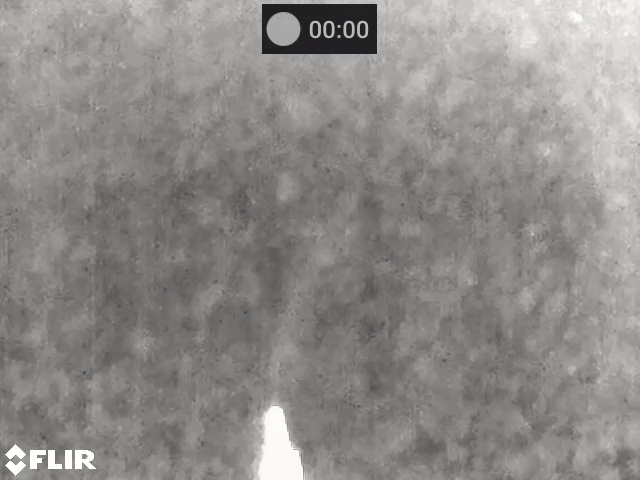}} \hspace{-2pt}
    \setcounter{subfigure}{0}
    \subfloat[RGB]{\label{fig:dataset_patches_a}\includegraphics[width=0.25\columnwidth]{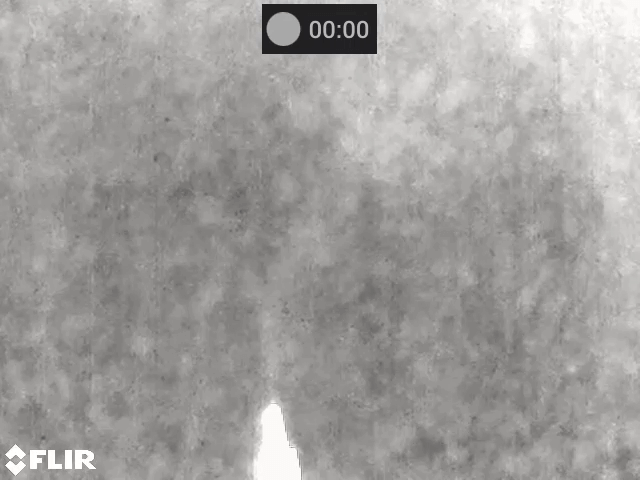}} \hspace{0pt}
    \subfloat{\includegraphics[width=0.25\columnwidth]{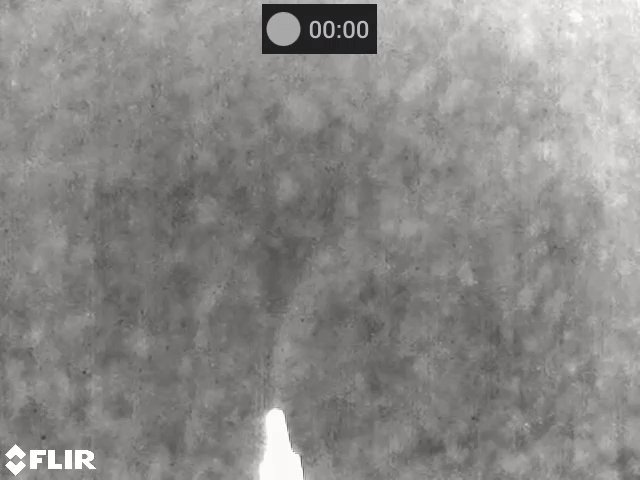}}  \\
   
    \setcounter{subfigure}{1}
    \subfloat{\includegraphics[width=0.25\columnwidth]{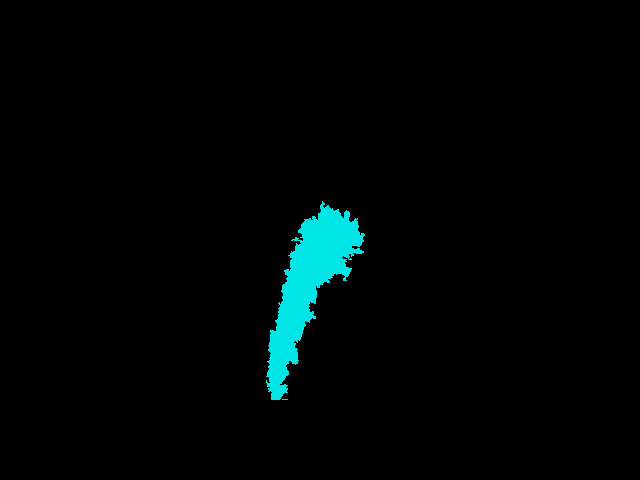}} \hspace{-2pt}
    \setcounter{subfigure}{1}
    \subfloat[Mask label]{\label{fig:dataset_patches_c}\includegraphics[width=0.25\columnwidth]{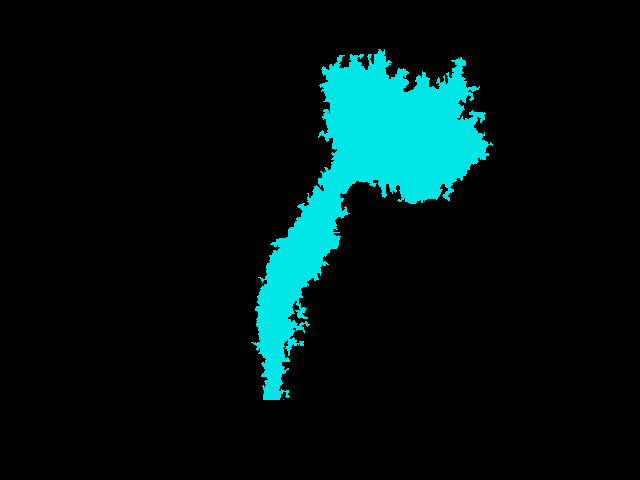}} \hspace{0pt}
    \subfloat{\includegraphics[width=0.25\columnwidth]{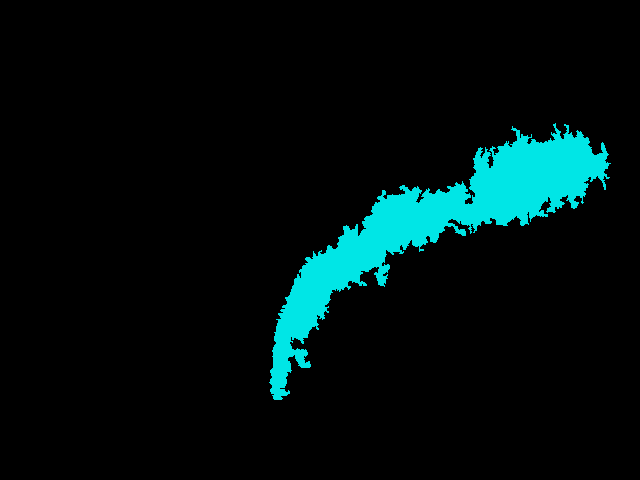}}  \\
    \vspace{-5pt}
    \caption{(a) CR dataset images. (b) Associated ground truth masks for gas plume regions.}
    \label{fig:dataset_rumen}
    \vspace{-5pt}
\end{figure}

\subsection{Controlled Methane Release Dataset}

To create the controlled methane release (MR) dataset, we set up an experiment where methane plumes are captured using the FLIR GF77 camera while controlling the flow rate of methane gas released from a Methane Calibration Gas Cylinder (UHP (99.995\%), UHP, 34 Litre Cylinder). A block of ice is used as the background to ensure a consistent temperature contrast between the background and the gas. The distance from the gas source to the background is fixed at 2 inches, and the distance from the gas source to the camera is set at 12 inches. The room temperature remains stable in the range of 27.8 - 28.8 degrees Celsius, and the gas absolute pressure (PSIA) is consistently recorded at 13.83 throughout the experiment. 

A precision gas flow controller (Cole-Parmer Digital Pressure Controller, 0 to 15 psi, 1/8" NPT(F)) connected to a methane gas tank is used to regulate the flow of methane. The flow rate is incrementally adjusted in 10 SCCM (Standard Cubic Centimeters per Minute) intervals, starting from 10 SCCM and reaching up to 100 SCCM. At each flow rate, methane gas is released, and videos of the resulting plume are captured using the GF77 camera. 

Prior to each gas release, a 15-second video is recorded without any gas present to serve as a background for subsequent subtraction. This step helps in isolating the effects of the gas plume in the captured footage. In total, 15 videos with methane plumes are recorded from the GF77 camera, and frames are extracted from these videos to create the dataset. The MR dataset, outlined in \Cref{tab:methane_dataset}, consists of images captured at different flow rates and using different color modes, as shown in \cref{fig:dataset_methane}. This dataset is split into training (80\%), validation (10\%), and testing (10\%) subsets. In total, 9,237 images are extracted from the videos, each with a resolution of 640 x 480 pixels.

\begin{table}
  \centering
  \small
  \begin{tabular}{@{}ccccc@{}}
    \toprule
    Class (SCCM)  & Train & Test & Val & Color Mode \\
    \midrule
    10 & 282 & 35 & 36 & White hot\\
    20 & 967 & 120 & 123 & White hot, Black hot\\
    30 & 618 & 77 & 78 & White hot\\
    40 & 722 & 90 & 91 & White hot \\
    50 & 889 & 111 & 112 & White hot\\
    60 & 927 & 115 & 119 & White hot, Black hot\\
    70 & 1215 & 200 & 199 & White hot \\
    80 & 693 & 85 & 89 & White hot, Rainbow \\
    90 & 574 & 71 & 73 & White hot\\
    100 & 420 & 52 & 54 & White hot \\
    \bottomrule
  \end{tabular}
  \vspace{-5pt}
  \caption{Controlled methane release (MR) dataset distribution across flow rate classes (SCCM), with training, testing, and validation splits. The Color Mode column shows the color representations used by the FLIR GF77 camera for visualizing gas plumes.}
  \label{tab:methane_dataset}
  \vspace{-5pt}
\end{table}


\begin{table}
  \centering
  \small
  \begin{tabular}{@{}ccccc@{}}
    \toprule
     & Train & Test & Validation  & Color Mode \\
    \midrule
    & 272 & 34 & 34  & White hot \\
    \bottomrule
  \end{tabular}
  \vspace{-5pt}
  \caption{Dairy cow rumen (CR) dataset distribution, with training, testing, and validation splits.}
  \label{tab:rumen_dataset}
  \vspace{-5pt}
\end{table}

\subsection{Dairy Cow Rumen Gas Dataset}
The dairy cow rumen gas (CR) dataset is created using the method mentioned in the controlled methane release dataset, with a few key differences. Instead of using a gas cylinder to release methane, dairy cow rumen gas samples are collected from a farm in TEDLAR gas bags (CEL Scientific Corp., Santa Fe Springs, CA, USA). These gas samples are obtained using a 24-hour batch culture system to measure the total gas and methane production from dairy cow rumen samples. The ANKOM gas production system (ANKOM Technology, Macedon, NY) is used for the incubation, where 100 ml of rumen liquor collected from a Holstein dairy cow is incubated with 100 ml of buffer \cite{McDougall1948-pi} and 3 grams of a total mixed ration (TMR) consisting of 50\% alfalfa hay and 50\% concentrate mix. The ANKOM modules are flushed with CO2 to maintain anaerobic conditions and then moved to a 39\textdegree C water bath to maintain the optimal temperature for the bacteria. Following the incubation, TEDLAR gas bags are connected to the modules to collect the emitted gas required for methane analysis. After 24 hours of incubation, all gas bags are collected and analyzed.

The collected dairy cow rumen gas samples are then used to release the gas and capture methane plume video using the FLIR GF77 camera, as shown in \cref{fig:dataset_rumen}. For this dataset, a single video is recorded from the gas bag, comprising 340 images with a resolution of 640 x 480 pixels. These images are divided into training, testing, and validation sets, following an 80\%, 10\%, and 10\% split ratio, respectively. \Cref{tab:rumen_dataset} presents an overview of the training, testing, and validation sets, along with the color mode used during video capture on the CR dataset.



\begin{table}
\centering
\setlength{\tabcolsep}{2pt}
\resizebox{0.95\columnwidth}{!}{%
\begin{tabular}{l c c c c c c}
\toprule
\multicolumn{2}{c}{\multirow{2}{*}{Hyperparameters}} &
\multicolumn{5}{c}{Controlled Methane Release Dataset} \\ 
\cmidrule(r){3-7}
& & mIoU &
mFscore &
FPS &
GFlOPs &
Params. (M)  \\ 

\midrule

\multicolumn{1}{c}{
    \multirow{2}{*}{
        \begin{tabular}[c]{@{}c@{}}
            Encoder \\ Features
        \end{tabular}
    }
} & 
3 Stage & 
\multicolumn{1}{c}{85.1} &   
91.73 &
\multicolumn{1}{c}{110.66} & 
4.88 & 
3.643    \\ 

\multicolumn{1}{c}{} & 
4 Stage & 
\multicolumn{1}{c}{85.9} &
92.19 &
\multicolumn{1}{c}{97.45} &
9.951 & 
3.652 \\

\midrule

\multicolumn{1}{c}{
    \multirow{3}{*}{
        \begin{tabular}[c]{@{}c@{}}
            Channel \\ Dimension
        \end{tabular}
    }
} &
128 & 
85.76 &
92.1 & 
108.98 & 
5.836 & 
3.436 \\ 

\multicolumn{1}{c}{} & 
256       & 
85.9 &
92.19 & 
\multicolumn{1}{c}{97.45} &   
9.951 & 
3.652 \\ 

\multicolumn{1}{c}{} & 
512 & 
85.93 &
92.23 & 
81.53 &
23.013 &  
4.377 \\ 
\bottomrule
\end{tabular}}
\vspace{-5pt}
\caption{Ablation study on the design of Gasformer. 3 stage means the encoder's output of stages 2, 3 and 4 is sent to the decoder. 4 stage means the encoder's output of stages 1, 2, 3, and 4 is sent to the decoder. GFLOPs is calculated with the input size of 512 x 512.}

\label{tab:abla}
\vspace{-5pt}
\end{table}


\begin{table}
  \centering
  \small
  \begin{tabular}{@{}lccc@{}}
    \toprule
    Architecture & Image Size & mIoU & mFscore \\
    \midrule
    FCN & 512 x 512 & 57.54 & 71.08 \\
    DeepLabv3$+$ & 512 x 512 & 57.78 & 69.8 \\
    SegNeXt-T & 512 x 512 & 83.75 & 90.89 \\
    Segformer B0 & 512 x 512 & 85.37 & 91.94 \\
    Gasformer (ours) & 512 x 512 & \textbf{85.9} & \textbf{92.19} \\
    \bottomrule
  \end{tabular}
  \vspace{-5pt}
  \caption{Comparison with existing methods on MR dataset. Ours is better.}
  \label{tab:methane_result}
  \vspace{-5pt}
\end{table}


\begin{table*}[!t]
\vspace{-0.4cm}
\centering
\setlength{\tabcolsep}{2pt}
\resizebox{\textwidth}{!}{
\renewcommand{\arraystretch}{1.3}
\begin{tabular}{l  c c  c c  c c  c c  c c  c  c  c c  c c  c c  c c  c c}
\toprule

\multirow{2}{*}{Model} &
\multicolumn{11}{c}{IoU} &
\multicolumn{11}{c}{Fscore} \\
\cmidrule(r){2-12}
\cmidrule(r){13-23}

& Background & 10 & 20 & 30 & 40 & 50 & 60 & 70 & 80 & 90 & 100 & Background & 10 & 20 & 30 & 40 & 50 & 60 & 70 & 80 & 90 & 100\\

\midrule
FCN	&	98.04 &	65.74 &	64.05 &	72.97 &	53.23 &	45.7 &	53.31 &	71.45 &	19.95 &	45.15 &	43.35 & 99.01 &	79.33 &	78.09 &	84.37 &	69.47 &	62.73 &	69.55 &	83.35 &	33.27 &	62.21 &	60.48 \\

DeepLabv3$+$		 & 98.16	 & 67.34	 & 77.23 & 	70.84	 & 48.6	 & 34.53 & 	71.02 & 	77.96	 & 20.55 & 	54.11 & 	15.24 & 99.07 & 	80.49	 & 87.15	 & 82.93 & 	65.41 & 	51.33 & 	83.05 & 	87.62 & 	34.1	 & 70.22 & 	26.46\\

SegNext-T	 & 	99.22 & 	61.37	 & 75.22 & 	84.45	 & 84.66	 & 84.27 & 	84.29	 & 89.21	 & 85.76 & 	87.1	 & 85.76 & 99.61 & 	76.06 & 	85.86 & 	91.57 & 	91.69 & 	91.47 & 	91.48	 & 94.3 & 	92.33 & 	93.1 & 	92.33\\

Segformer-B0	&	99.29	 & \textbf{67.83} & 	80.33 & 	84.39 & 	84.96	 & 84.91	 & 86.92 & 	89.53	 & 88.72	 & 87.59	 & 84.63 & 99.64 & 	\textbf{80.83} & 	89.09 & 	91.53 & 	91.87	 & 91.84	 & 93 & 94.47	 & 94.02	 & 93.38	 & 91.68 \\

Gasformer	 & 	\textbf{99.34}	 & 64 & 	\textbf{81.7} & 	\textbf{85.01}	 & \textbf{86.01} & 	\textbf{85.87} & 	\textbf{87.44} & 	\textbf{90.08} & 	\textbf{90.39} & 	\textbf{88.58}	 & \textbf{86.47} & \textbf{99.67}	 & 78.05	 & \textbf{89.93}	 & \textbf{91.9} & 	\textbf{92.48}	 & \textbf{92.4} & 	\textbf{93.3} & 	\textbf{94.78}	 & \textbf{94.95} & 	\textbf{93.94} & 	\textbf{92.74} \\

\bottomrule
\end{tabular}}
\vspace{-5pt}
\caption{Per class result on MR dataset. Class labels are SCCM, except for the background class.}
\label{table:methane_classwise_result}
\vspace{-5pt}
\end{table*}

\subsection{Mask generation}


To generate labeled data for semantic segmentation tasks, we employ a semi-automated data labeling process.
The process begins with background subtraction to isolate methane plumes from foreground images by computing an average image from background images collected prior to each gas release and subtracting it from each foreground image captured after the gas release.
The contrast of the resulting image is then enhanced to facilitate better object segmentation. An adaptive thresholding technique is then applied to convert the preprocessed images into a binary format, distinctly separating the methane plumes from the background. The adaptive thresholding parameters need to be manually tuned for each gas concentration based on factors such as camera color mode and gas plume shape to obtain good segmentation masks.

The watershed algorithm \cite{watershed}, guided by an elevation map generated using a Sobel filter, is used to refine the segmentation and accurately delineate the boundaries of the methane plumes. Region properties analysis is then conducted to identify and examine each segmented object, selecting those that surpass a predefined size threshold for annotation. This step filters out insignificant objects and ensures that only gas plumes are considered for mask generation. The size threshold for region properties analysis has to be adjusted based on the shape and size of the gas plume in each scenario to accurately identify the methane regions.

For the CR dataset specifically, manual intervention is required to specify separation lines to isolate the methane plume from the connected gas release nozzle of the gas bags. During the mask generation process, when adaptive thresholding is applied, the gas release nozzle is included in the initial mask along with the methane plume. To address this issue, we manually set y-coordinate thresholds to separate the methane plume from the gas release nozzle, effectively removing the nozzle from the final mask.

The manual tuning process for adaptive thresholding and region size thresholding is repeated across color modes, gas concentrations, plume shapes, and datasets, while the separation line specification is only necessary for the CR dataset. 
Finally, binary masks are generated for each identified object, providing pixel-wise annotations crucial for semantic segmentation tasks. These masks are saved as separate images, with each mask corresponding to a specific plume, as shown in \cref{fig:dataset_methane} and \cref{fig:dataset_rumen}.

\section{Experiments and Discussion}

\subsection{Implementation Details}
We implement our experiments using the Pytorch and MMsegmentation \cite{mmseg2020} framework. The experiments are performed on a computer with an Intel Core i9 11900F CPU running at 2.50GHz, 32 GB of memory, and an NVIDIA RTX 3090 graphics card (10,496 CUDA cores, 24 GB dedicated graphics memory). The encoders of our segmentation models are pre-trained on the ImageNet-1K dataset. We evaluate the models using mean Fscore (mFscore) for classification tasks and mean Intersection over Union (mIoU) for segmentation tasks. All models use the AdamW optimizer with a learning rate of 0.00006, beta coefficients of (0.9, 0.999), and a weight decay of 0.01. The learning rate schedule consists of a LinearLR warm-up phase for the first 1500 iterations, followed by a PolynomialLR decay with a power of 1.0. We train for 160,000 iterations, validating every 16,000 iterations, using a batch size of 2 for training and 1 for validation and testing. We adopt several data augmentation steps during the training phase to enhance model learning and generalization. Initially, we randomly resize images and their corresponding segmentation masks to 640 x 480 within a ratio range of 0.5 to 2, maintaining their aspect ratio. Following resizing, we randomly crop images into 512 x 512, introducing spatial variability.


\subsection{Ablation Study}
In our ablation study, we investigate the impact of encoder features and decoder channel dimension on the performance of Gasformer.

\noindent\textbf{Encoder Features.}
When comparing the use of 3 stages (stage 2, stage 3, stage 4) and 4 stages (all stages) of multi-level features from the MiT B0 encoder, as shown in \cref{tab:abla}, we find that using 4 stages results in a 0.94\% improvement in mIoU and a 0.5\% improvement in mFscore. However, this comes at the cost of a decrease of 13.2 FPS in inference speed and with only 1.003x more parameters.

\noindent\textbf{Decoder Channel Dimension.}
When we explore the effect of the decoder channel dimension, as shown in \cref{tab:abla}, we observe that increasing it from 128 to 256 leads to a 0.16\% improvement in mIoU and a 0.09\% improvement in mFscore. Further increasing the channel dimension to 512 results in a 0.03\% mIoU improvement and a 0.04\% mFscore improvement compared to the 256-channel configuration. We also note that the FPS decreases significantly with higher channel dimensions, with the 256-channel configuration being 1.12x slower and the 512-channel configuration being 1.34x slower compared to the 128-channel configuration. Additionally, the GPU memory usage and the number of parameters increase with higher channel dimensions, with the 512-channel configuration being 1.27x larger in terms of parameters and 3.94x more computationally expensive in terms of FLOPs compared to the 128-channel configuration.

Based on our findings, we chose to use 4 stages of encoder features and a decoder channel dimension of 256 to achieve an optimal balance between segmentation performance and computational efficiency.

\subsection{Results}
\subsubsection{Segmentation Results on the MR Dataset}

In the quantitative analysis presented in \cref{tab:methane_result}, the proposed method, Gasformer, is compared with other state-of-the-art segmentation methods, including FCN \cite{fcn}, DeepLabV3+ \cite{deeplabv3plus}, SegNext \cite{Guo2022-bj}, and Segformer \cite{Xie2021-in}. 
The results show that the proposed Gasformer method outperforms the other methods in all metrics. Specifically, Gasformer achieves the highest mIoU and mFscore scores of 85.9 and 92.19, respectively. Segformer follows closely with an mIoU of 85.37 and mFscore of 91.94.

\begin{figure*}
    \centering

    \captionsetup[subfloat]{labelformat=empty}


    \rotatebox[origin=c]{90}{\parbox{0cm}{\centering 40}}\hspace{5pt}%
    \subfloat{\includegraphics[width=0.28\columnwidth]{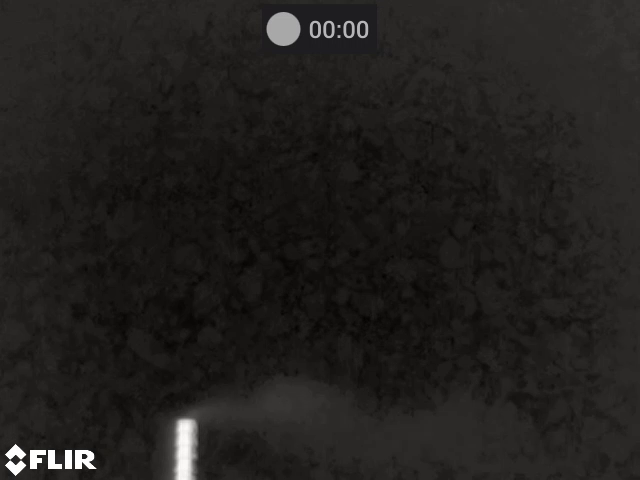}} \hspace{0pt}
    \subfloat{\includegraphics[width=0.28\columnwidth]{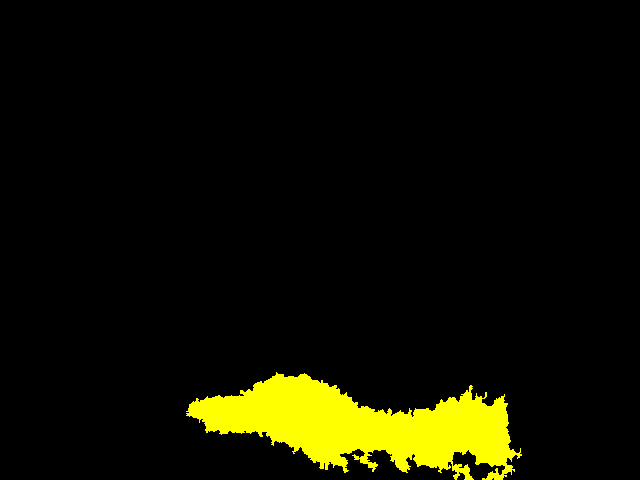}} \hspace{0pt}
    \subfloat{\includegraphics[width=0.28\columnwidth]{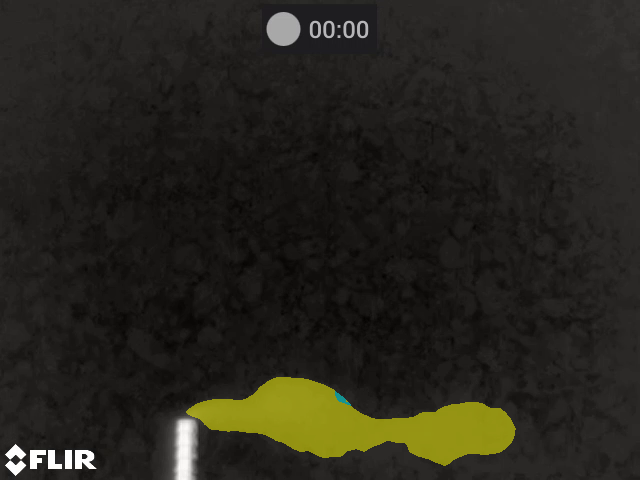}} \hspace{0pt}
    \subfloat{\includegraphics[width=0.28\columnwidth]{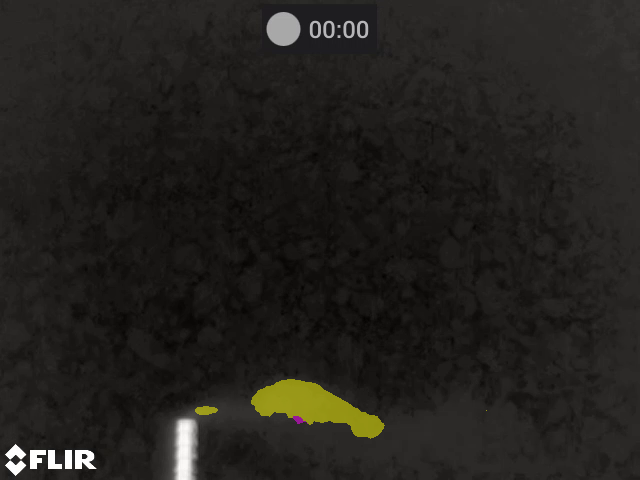}} \hspace{0pt}
    \subfloat{\includegraphics[width=0.28\columnwidth]{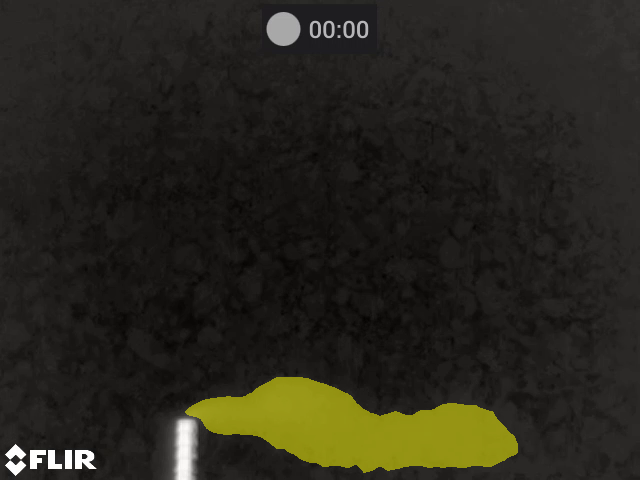}} \hspace{0pt}
    \subfloat{\includegraphics[width=0.28\columnwidth]{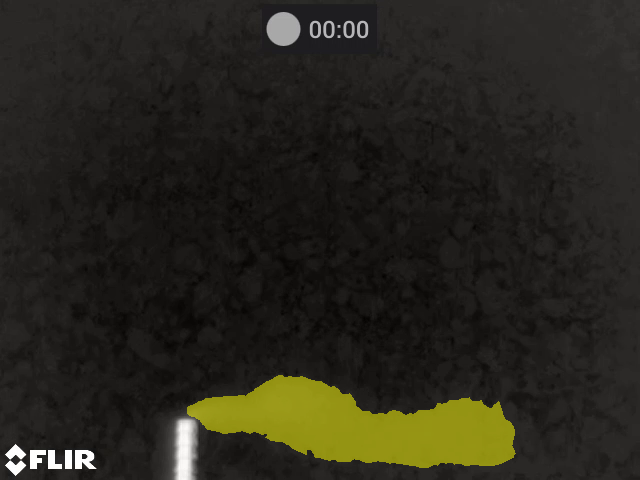}} \hspace{0pt}
    \subfloat{\includegraphics[width=0.28\columnwidth]{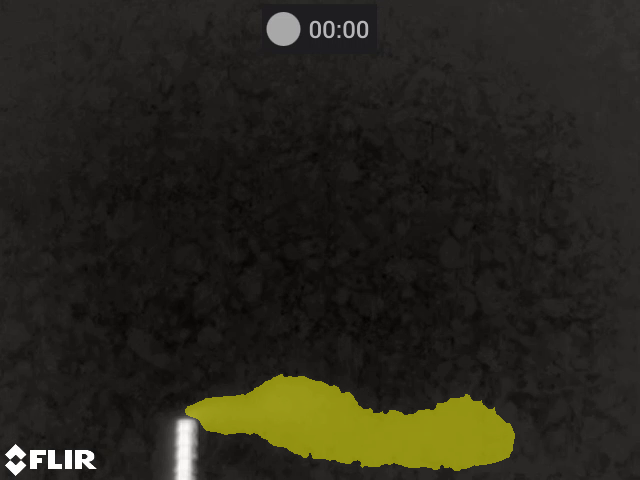}}  \\
    \rotatebox[origin=c]{90}{\parbox{0cm}{\centering 50}}\hspace{5pt}%
    \subfloat{\includegraphics[width=0.28\columnwidth]{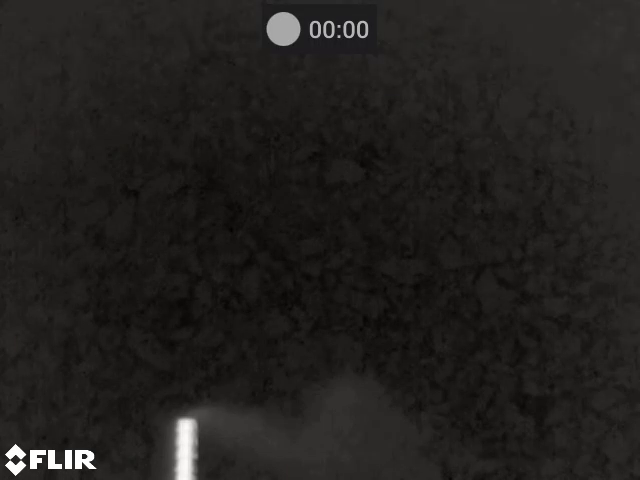}} \hspace{0pt}
    \subfloat{\includegraphics[width=0.28\columnwidth]{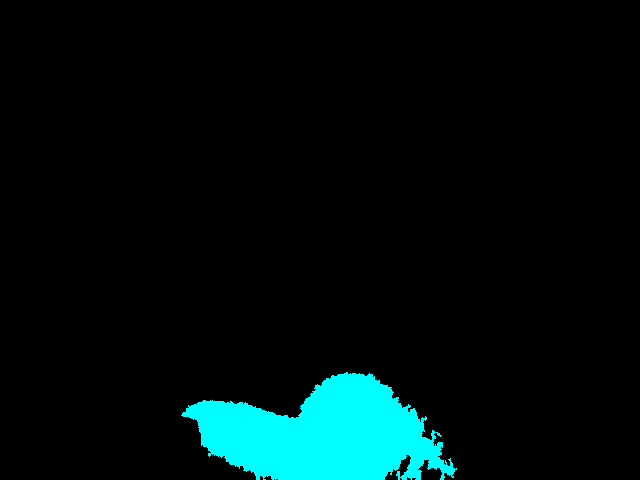}} \hspace{0pt}
    \subfloat{\includegraphics[width=0.28\columnwidth]{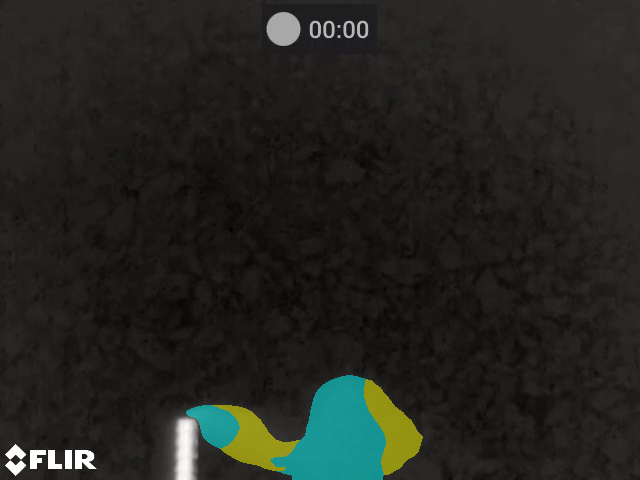}} \hspace{0pt}
    \subfloat{\includegraphics[width=0.28\columnwidth]{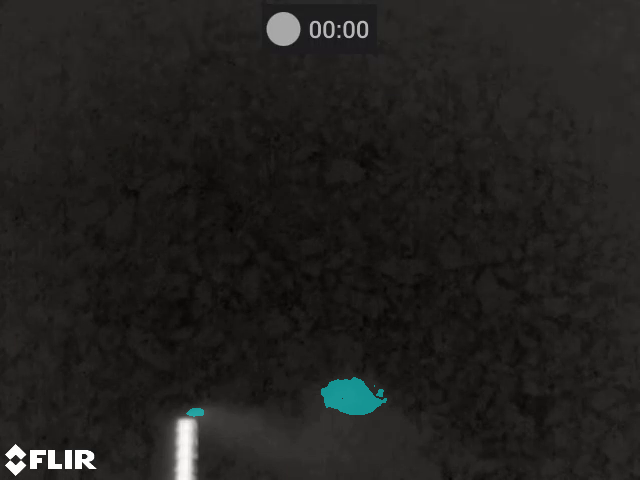}} \hspace{0pt}
    \subfloat{\includegraphics[width=0.28\columnwidth]{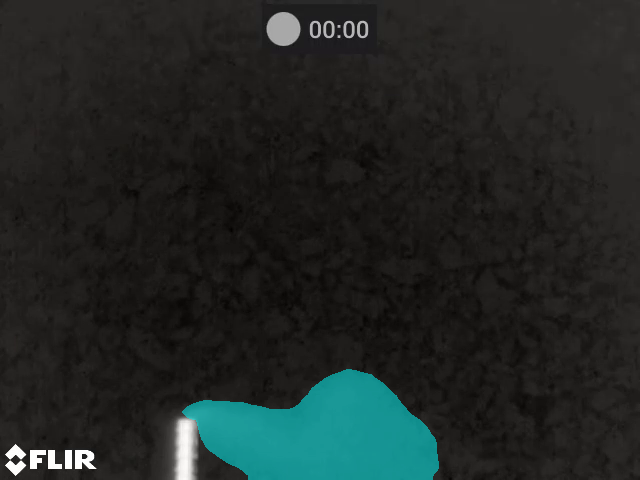}} \hspace{0pt}
    \subfloat{\includegraphics[width=0.28\columnwidth]{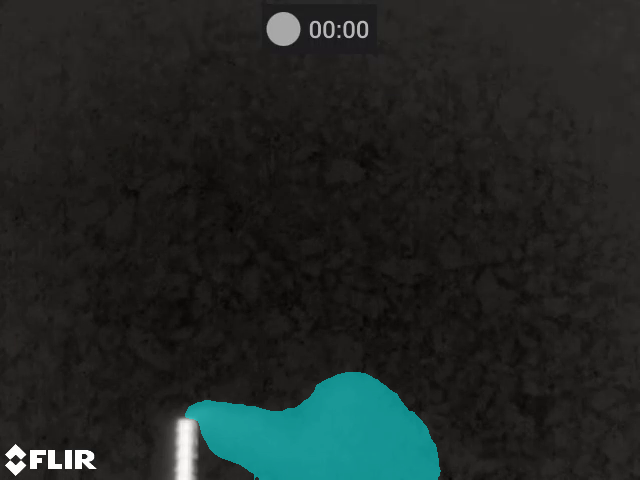}} \hspace{0pt}
    \subfloat{\includegraphics[width=0.28\columnwidth]{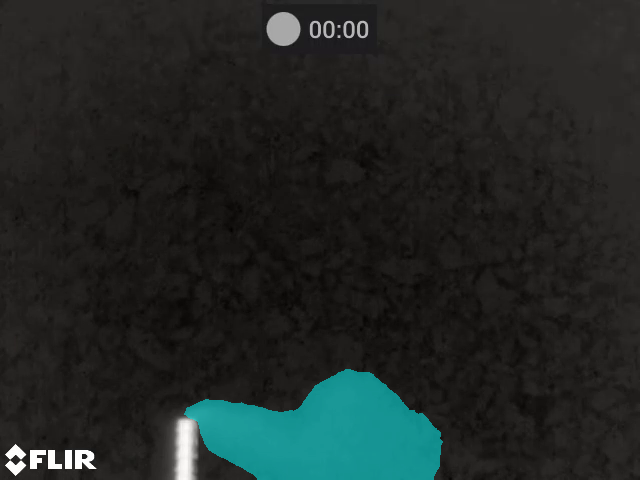}}  \\

    \rotatebox[origin=c]{90}{\parbox{0cm}{\centering 80}}\hspace{5pt}%
    \subfloat[Image]{\includegraphics[width=0.28\columnwidth]{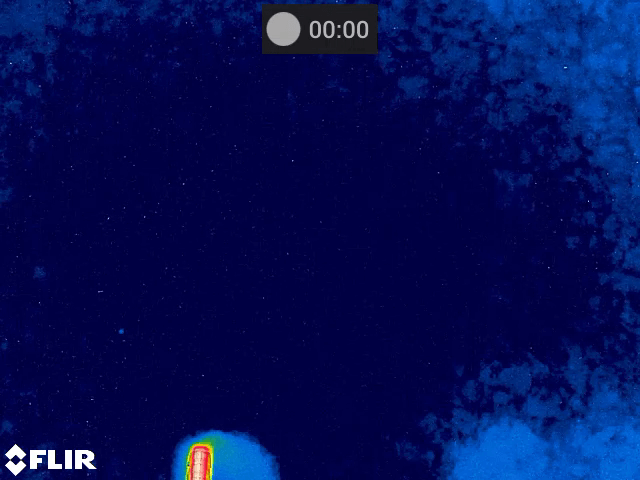}} \hspace{0pt}
    \subfloat[Ground Truth]{\includegraphics[width=0.28\columnwidth]{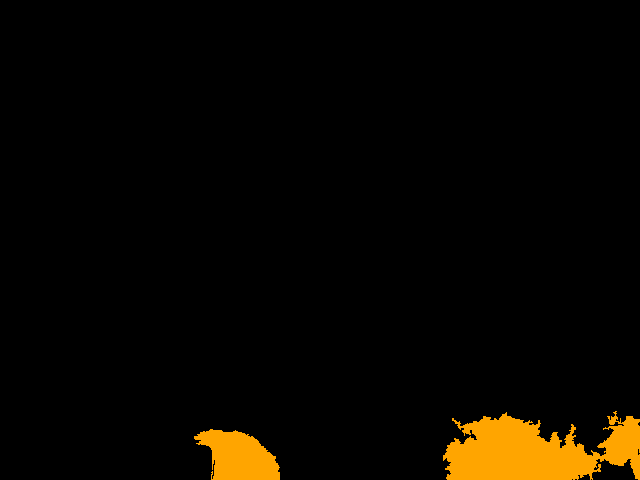}} \hspace{0pt}
    \subfloat[FCN]{\includegraphics[width=0.28\columnwidth]{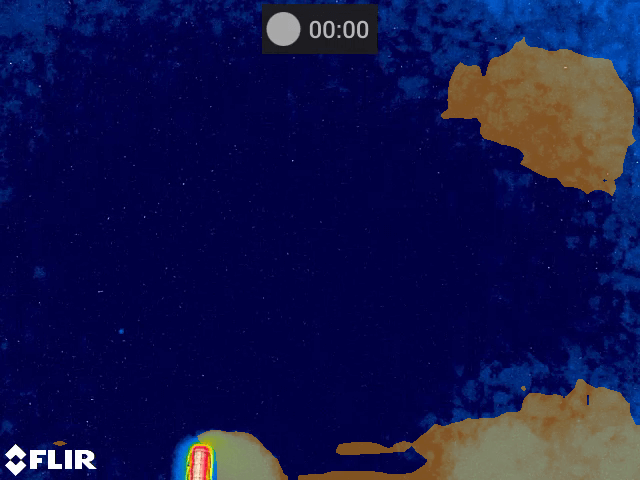}} \hspace{0pt}
    \subfloat[DeepLabV3+]{\includegraphics[width=0.28\columnwidth]{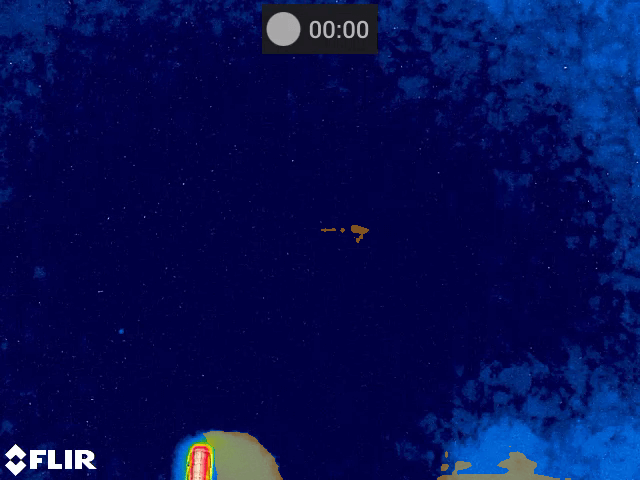}} \hspace{0pt}
    \subfloat[SegNeXt]{\includegraphics[width=0.28\columnwidth]{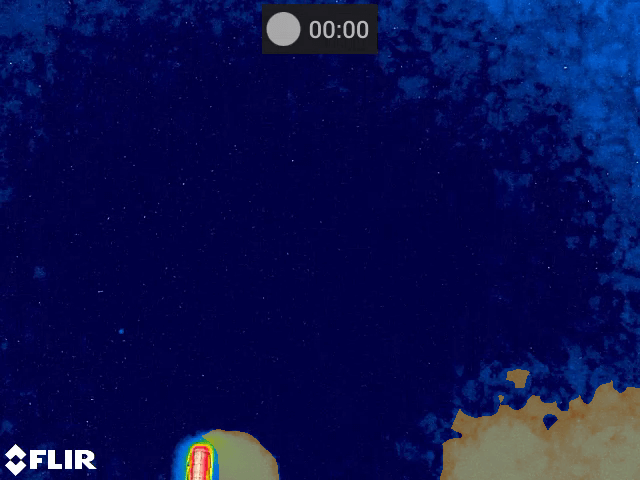}} \hspace{0pt}
    \subfloat[Segformer]{\includegraphics[width=0.28\columnwidth]{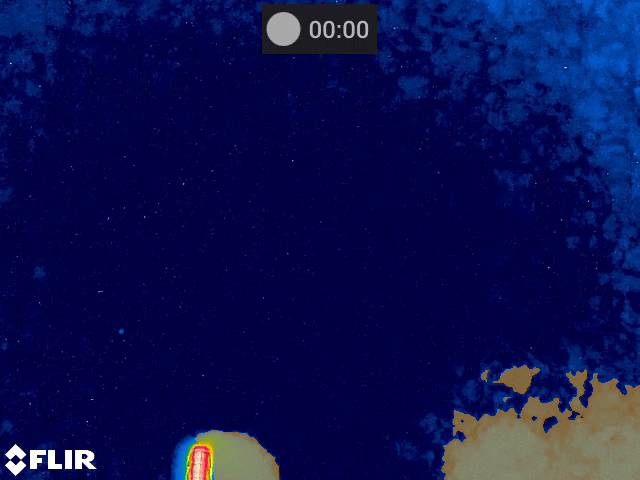}} \hspace{0pt}
    \subfloat[Gasformer]{\includegraphics[width=0.28\columnwidth]{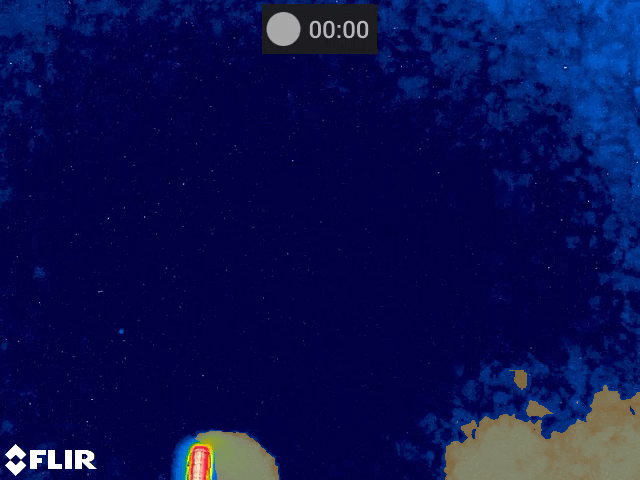}}  \\
    \vspace{-5pt}
    \caption{Segmentation performance of different models on MR dataset, showing the input images, ground truth masks, and the segmentation outputs for 40 SCCM, 50 SCCM, and 80 SCCM methane flow rates.}
    \label{fig:methane_predictions}
    \vspace{-5pt}
\end{figure*}

It is crucial to analyze the performance of the methods at different gas flow rates presented in \cref{table:methane_classwise_result}, especially at low flow rates where the background and gas have very low contrast. The proposed method demonstrates superior performance in both IoU and Fscore for low flow rates ranging from 20 SCCM to 100 SCCM, except for the 10 SCCM case where Segformer achieves the best result.
Additionally, the proposed method excels in segmenting the background, followed by Segformer, SegNeXt, DeepLabv3+, and FCN.

\begin{table}
  \centering
  \small
  
  \begin{tabular}{@{}lccc@{}}
    \toprule
    Architecture & Image Size & mIoU & mFscore \\
    \midrule
    FCN & 512 x 512 & 86.41 & 92.24 \\
    DeepLabv3+ & 512 x 512 & 88.18 & 93.37 \\
    SegNeXt-T & 512 x 512 & 88 & 93.26 \\
    Segformer B0 & 512 x 512 & 88.41 & 93.52 \\
    Gasformer (ours) & 512 x 512 & \textbf{88.56} & \textbf{93.61} \\
    \bottomrule
  \end{tabular}
  \vspace{-5pt}
  \caption{Comparison with existing methods on CR dataset. Ours is better.}
  \label{tab:rumen_result}
  \vspace{-5pt}
\end{table}

\begin{table}[!t]
\centering
\small
\resizebox{0.95\columnwidth}{!}{%
\begin{tabular}{l  c c  c c  c c  c c  c c  c  c  c c  c c  c c  c c  c c}
\toprule

\multirow{2}{*}{Model} &
\multicolumn{2}{c}{IoU} &
\multicolumn{2}{c}{Fscore} \\
\cmidrule(r){2-3}
\cmidrule(r){4-5}

& Background & Gas & Background & Gas\\
\midrule
FCN &	98.7 &	74.11	 & 99.35 & 	85.13 \\

DeepLabv3+ & 98.87 & 77.49 & 	99.43 & 87.32 \\

SegNext-T	 & 98.85 & 	77.16	 & 99.42 & 	87.1 \\

Segformer-B0  & 	98.89	 & 77.94 & 	99.44 & 	87.6 \\ 

Gasformer	 & 	\textbf{98.9} & 	\textbf{78.22}	 & \textbf{99.45} & 	\textbf{87.78} \\

\bottomrule
\end{tabular}}
\vspace{-5pt}
\caption{Per class results of CR dataset.}
\label{table:rumen_classwise_result}
\vspace{-5pt}
\end{table}

\subsubsection{Segmentation Results on the CR Dataset}

We employ a transfer learning approach to train the segmentation models to address the limited number of images in the CR dataset, which contains only 272 images for training. The process involves pre-training the models on the larger MR dataset. After pre-training, we fine-tuned the models using the weights obtained from the previous stage.

The comparative analysis of cow rumen gas segmentation is presented in \cref{tab:rumen_result}. Rumen gas segmentation is more challenging than controlled low-flow rate methane segmentation due to the low contrast between the gas and the background. Our proposed method, Gasformer, achieves the best result, with a mIoU of 88.56 and mFscore of 93.61, followed by Segformer B0, DeepLabv3+, SegNeXt-T, and FCN.

\Cref{table:rumen_classwise_result} presents the individual performance of each model on background and gas segmentation tasks. Gasformer obtains the best results for both tasks, with an IoU of 98.9 and 78.22 for background and gas segmentation, respectively. Surprisingly, FCN, which performs the worst in the controlled flow rate methane task, achieves a mIoU of 86.41 in the cow rumen gas segmentation task. This improvement demonstrates the effectiveness of transfer learning using the controlled flow rate methane dataset.

\begin{table}
  \centering
  \small
  \begin{tabular}{@{}lcccc@{}}
    \toprule
    Architecture & FPS & GFLOPs & Params. (M) \\
    \midrule
    FCN & 47.1 & 198 & 47.129 \\
    DeepLabv3+ & 43.65 &	177	& 41.221\\
    SegNext-T & 84.33 & \textbf{6.305} & 4.228 \\
    Segformer B0 & \textbf{126.5} & 7.923 & 3.718 \\
    Gasformer & 97.45 & 9.951 & \textbf{3.652} \\
    \bottomrule
  \end{tabular}
  \vspace{-5pt}
  \caption{Real-time speed analysis of segmentation models. GFLOPs is calculated with the input size of 512 x 512}
  \label{tab:speed_analysis}
  \vspace{-5pt}
\end{table}

\begin{figure*}
    \centering

    \captionsetup[subfloat]{labelformat=empty}
    \subfloat{\includegraphics[width=0.28\columnwidth]{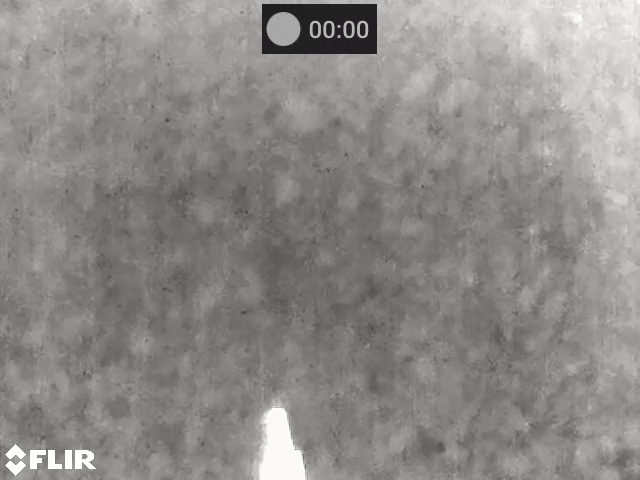}} \hspace{0pt}
    \subfloat{\includegraphics[width=0.28\columnwidth]{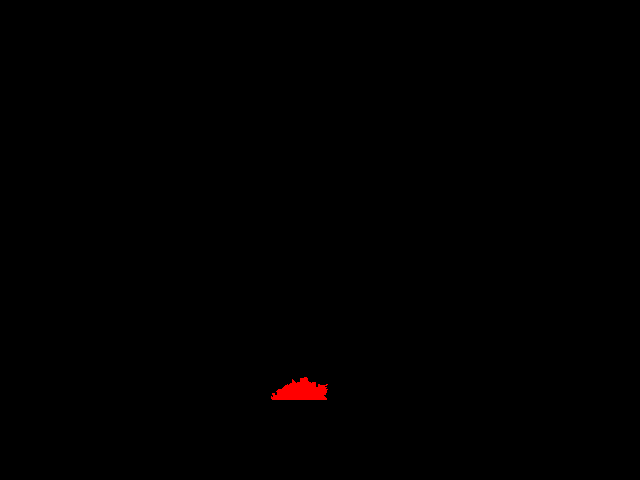}} \hspace{0pt}
    \subfloat{\includegraphics[width=0.28\columnwidth]{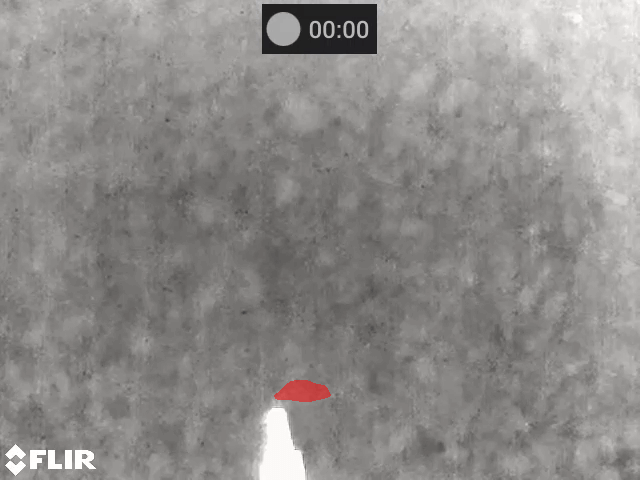}} \hspace{0pt}
    \subfloat{\includegraphics[width=0.28\columnwidth]{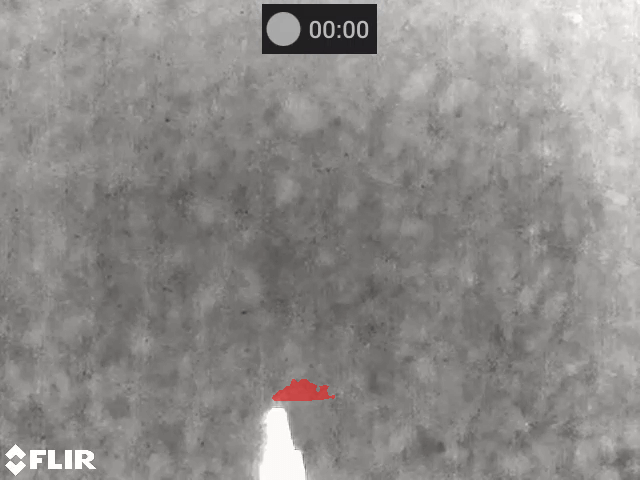}} \hspace{0pt}
    \subfloat{\includegraphics[width=0.28\columnwidth]{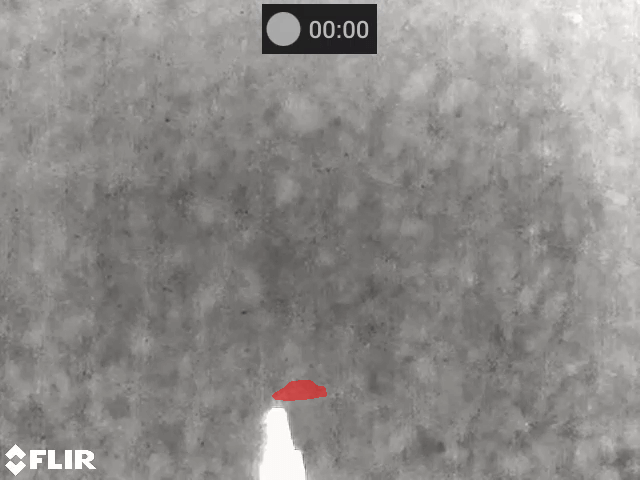}} \hspace{0pt}
    \subfloat{\includegraphics[width=0.28\columnwidth]{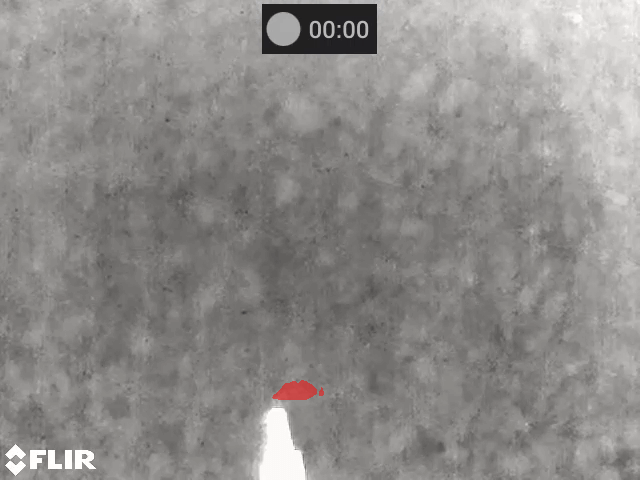}} \hspace{0pt}
    \subfloat{\includegraphics[width=0.28\columnwidth]{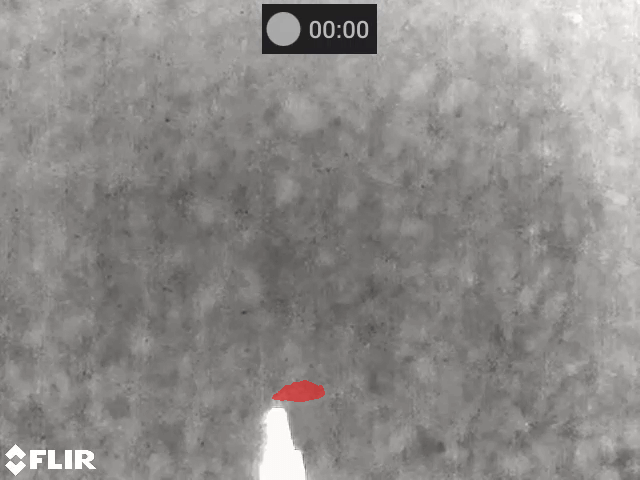}}  \\
    \subfloat{\includegraphics[width=0.28\columnwidth]{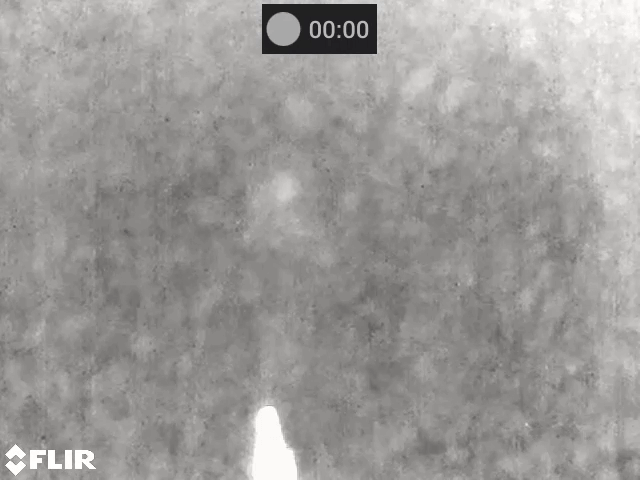}} \hspace{0pt}
    \subfloat{\includegraphics[width=0.28\columnwidth]{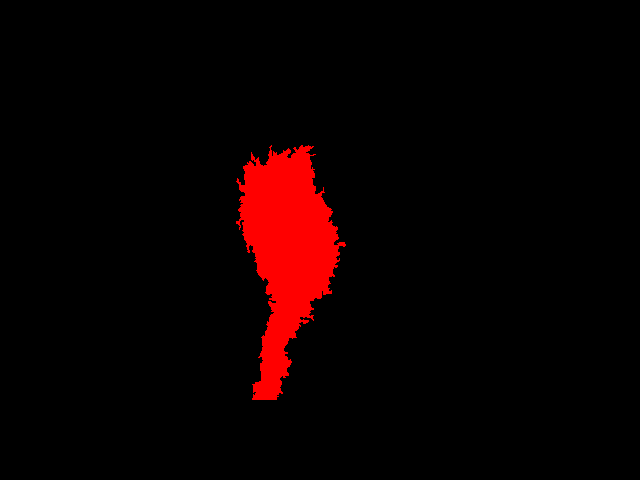}} \hspace{0pt}
    \subfloat{\includegraphics[width=0.28\columnwidth]{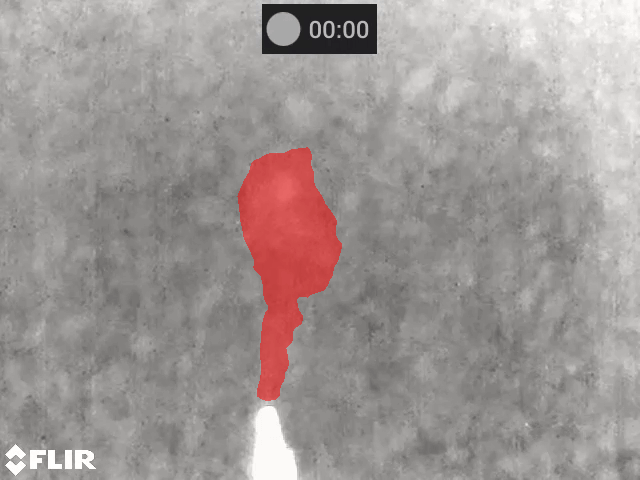}} \hspace{0pt}
    \subfloat{\includegraphics[width=0.28\columnwidth]{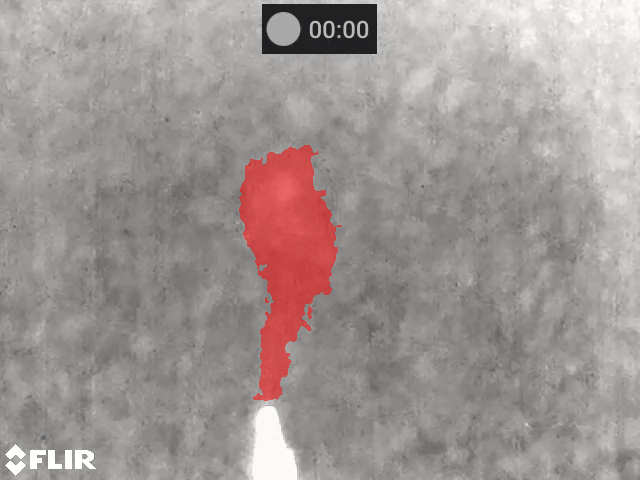}} \hspace{0pt}
    \subfloat{\includegraphics[width=0.28\columnwidth]{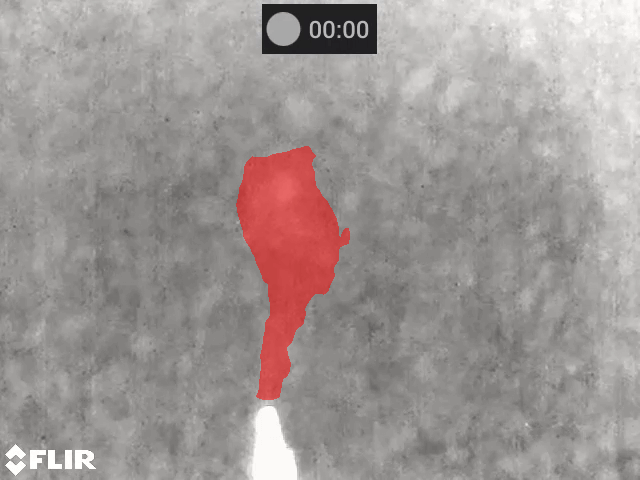}} \hspace{0pt}
    \subfloat{\includegraphics[width=0.28\columnwidth]{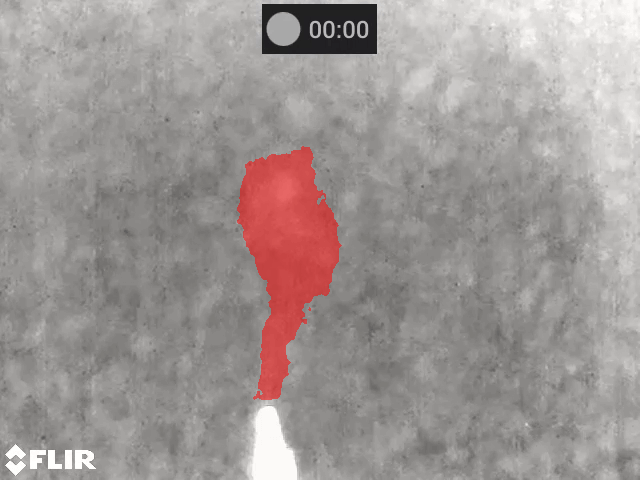}} \hspace{0pt}
    \subfloat{\includegraphics[width=0.28\columnwidth]{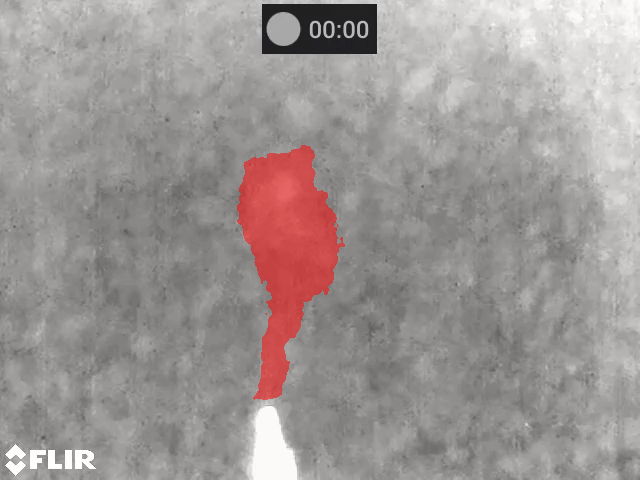}}  \\

    \subfloat[Image]{\includegraphics[width=0.28\columnwidth]{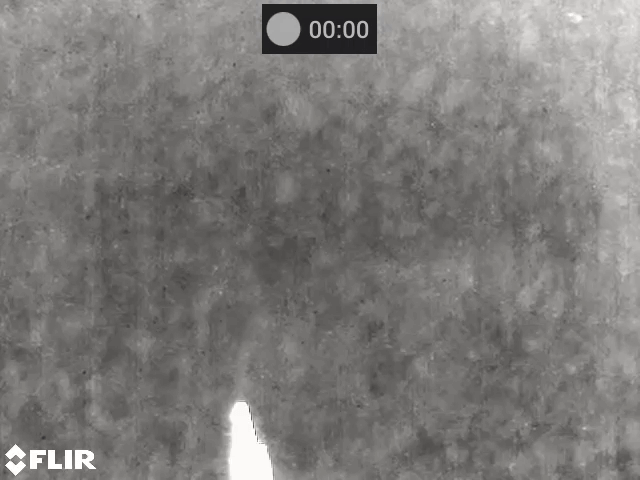}} \hspace{0pt}
    \subfloat[Ground Truth]{\includegraphics[width=0.28\columnwidth]{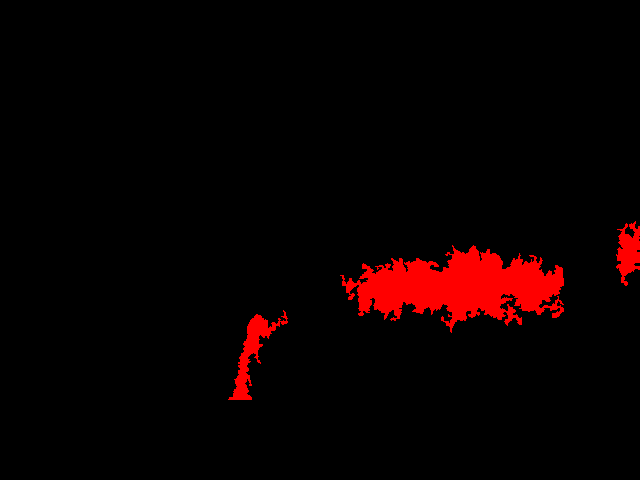}} \hspace{0pt}
    \subfloat[FCN]{\includegraphics[width=0.28\columnwidth]{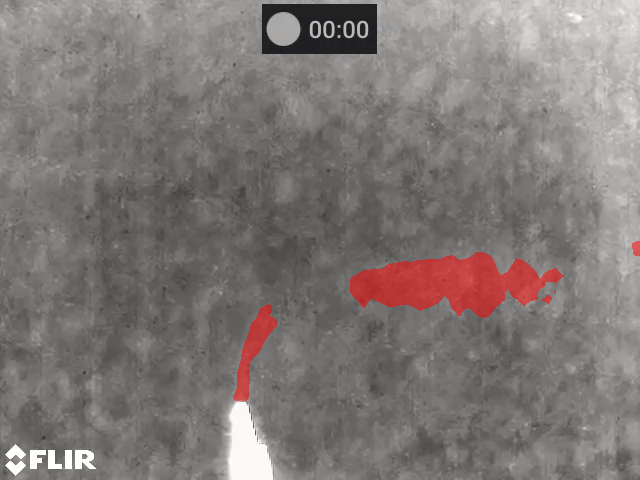}} \hspace{0pt}
    \subfloat[DeepLabV3+]{\includegraphics[width=0.28\columnwidth]{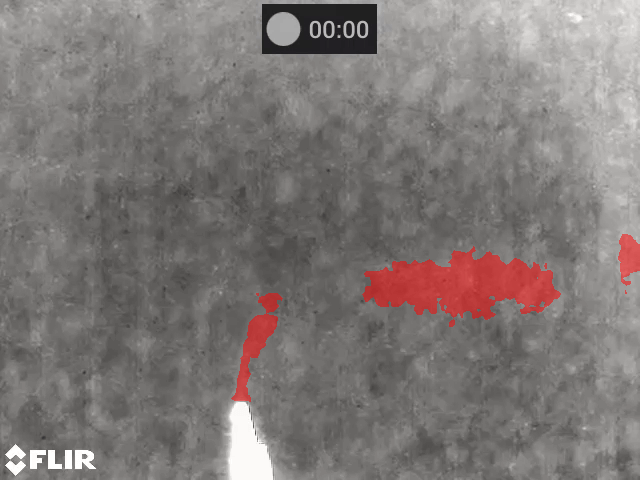}} \hspace{0pt}
    \subfloat[SegNeXt]{\includegraphics[width=0.28\columnwidth]{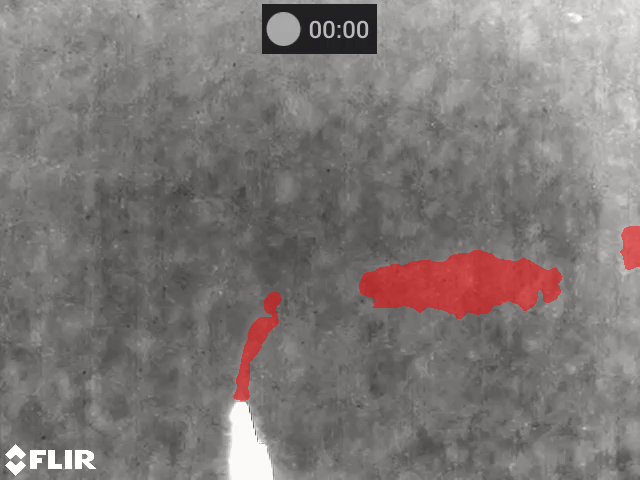}} \hspace{0pt}
    \subfloat[Segformer]{\includegraphics[width=0.28\columnwidth]{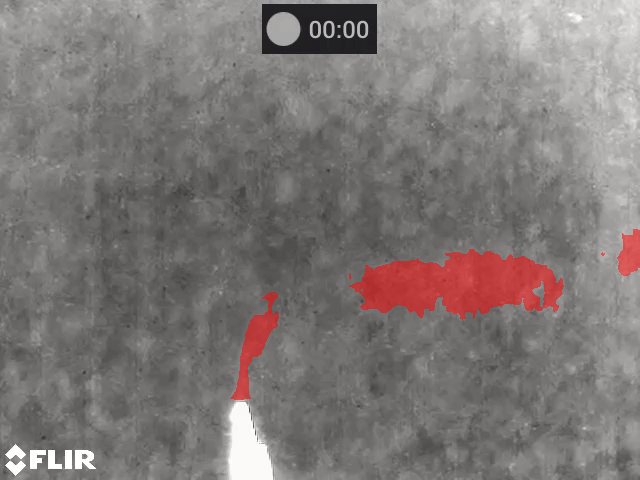}} \hspace{0pt}
    \subfloat[Gasformer]{\includegraphics[width=0.28\columnwidth]{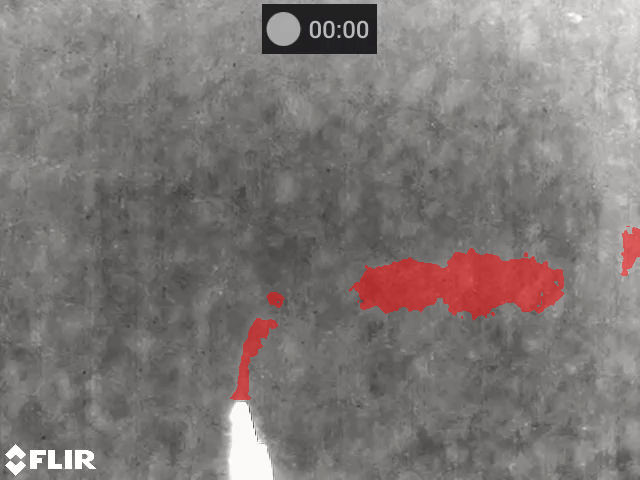}}  \\

    \vspace{-5pt}
    \caption{Semantic segmentation results of CR dataset.}
    \label{fig:rumen_predictions}
    \vspace{-5pt}
\end{figure*}


\subsubsection{Qualitative Comparison}
Qualitative examples of controlled methane gas release and dairy cow rumen gas segmentation are presented in \cref{fig:methane_predictions} and \cref{fig:rumen_predictions}, respectively. In \cref{fig:methane_predictions}, the second row shows a low flow rate of 50 SCCM, which has low contrast with the background. In this case, our proposed Gasformer method predicts satisfactory results and classifies the flow rate accurately. Although FCN was able to segment the gas, it failed to classify the flow rate correctly. Similarly, DeepLabV3+ struggled to segment the gas altogether. However, the proposed Gasformer method successfully segments the low-flow rate gas and classifies it correctly.

The main challenge arises in the CR dataset, as presented in \cref{fig:rumen_predictions}. The background visually resembles the gas, making it difficult to identify the gas emissions. Despite this similarity, the proposed Gasformer method demonstrates its ability to segment the gas accurately, even in the absence of sufficient visual cues for segmentation.

\subsubsection{Real-time Speed Analysis}
In the real-time time speed analysis presented in \cref{tab:speed_analysis}, the Gasformer architecture runs at 97.45 FPS on the GPU, which is 2.68x faster than the FCN architecture. However, it is 29.05 FPS slower than the Segformer B0 architecture, which achieves the highest FPS of 126.5.
In terms of FLOPS, Gasformer requires 1.57x more FLOPS compared to SegNext-T and 1.25x more FLOPS than Segformer B0.
Regarding model size, Gasformer is 12.9x smaller than FCN, 11.28x smaller than DeepLabv3+, 1.15x smaller than SegNext-T, and 1.01x smaller than Segformer B0.

\subsubsection{Sensor Impact on Gasformer's Performance}

As discussed in \cref{sec:flir}, the FLIR GF77 camera used in this study has a thermal sensitivity of $<$25 mK and a NECL gas sensitivity of at least 100 ppm-m for methane. The lower thermal sensitivity and higher NECL of the uncooled GF77 camera, compared to cooled cameras, can limit its effectiveness in detecting methane emissions at low flow rates or concentrations, making it more challenging to distinguish between the background and the methane plume when the temperature difference is small or the gas concentration is low.
To mitigate the impact of the camera's limitations, we designed our controlled release experiments to maintain a sufficient temperature contrast between the background and the methane plume, as well as various flow rates resulting in different methane concentrations. This approach allowed us to evaluate Gasformer's performance under diverse conditions, including scenarios that challenge the detection capabilities of the GF77 camera.

We included a representative range of flow rates in our dataset to provide a comprehensive assessment of Gasformer's performance, ensuring transparency in our methodology and results. It is important to note that the performance of Gasformer and other models evaluated in this study is inherently linked to the characteristics and limitations of the FLIR GF77 camera. The use of an uncooled LWIR camera may constrain the ability of any model to detect and segment methane emissions at very low flow rates or concentrations.


\vspace{-5pt}
\section{Conclusion}
In this study, we proposed Gasformer, a novel semantic segmentation architecture for detecting and quantifying methane emissions from livestock, and controlled release experiments using optical gas imaging. The architecture effectively captures multi-scale features and refines segmentation maps by combining a Mix Vision Transformer encoder and a Light-Ham decoder. We also introduced two unique datasets, the MR dataset, and the CR dataset, to evaluate the model's performance in controlled and real-world scenarios. Experimental results demonstrated that Gasformer outperforms state-of-the-art segmentation models on both datasets, achieving higher mIoU and mFscore values. The model's superior performance was particularly evident in low-contrast scenarios, such as low flow rates in the controlled dataset and challenging cow rumen gas segmentation tasks. The ablation study highlighted the importance of utilizing multi-scale features from the encoder and selecting an appropriate decoder channel dimension for optimal performance and computational efficiency.
Future research could explore the integration of Gasformer with other sensing modalities and cooled cameras to enhance its performance and adaptability in detecting various greenhouse gases and pollutants while investigating its robustness in challenging industrial and environmental settings.

\vspace{-5pt}
\section{Acknowledgment}

This work is supported by the National Institute of Food and Agriculture, U.S. Department of Agriculture, under award number 2022-70001-37404.
{
    \small
    \bibliographystyle{ieeenat_fullname}
    \bibliography{main}
}


\end{document}